%% file: MultiClassUSVM.tex
\pdfoutput=1

\documentclass{article}

\usepackage[nonatbib,final]{nips_2016}

\usepackage[utf8]{inputenc} 
\usepackage[T1]{fontenc}    
\usepackage{hyperref}       
\usepackage{url}            
\usepackage{booktabs}       
\usepackage{amsfonts}       
\usepackage{nicefrac}       
\usepackage{microtype}      
\usepackage{amsmath}
\usepackage{xfrac}
\usepackage[ruled,vlined]{algorithm2e}
\usepackage{amsthm}
\theoremstyle{plain}
\newtheorem{assumption}{Assumption}
\theoremstyle{definition}
\newtheorem{definition}{Definition}
\theoremstyle{plain}
\newtheorem{theorem}{Theorem}
\newtheorem{corollary}{Corollary}
\usepackage{bbm}
\usepackage{cite}
\usepackage{wrapfig}
\usepackage{xcolor}
\usepackage{enumitem}
\usepackage{subfigure}
\usepackage{array}
\usepackage{multirow}
\usepackage{pdfpages}
\newcommand{\specialcell}[2][c]{%
  \begin{tabular}[#1]{@{}c@{}}#2\end{tabular}}

\title{Universum Learning for Multiclass SVM}

\author{Sauptik Dhar$^{\dagger}$ \quad Naveen Ramakrishnan$^{\dagger}$ \quad \textbf{Vladimir Cherkassky}$^{\ddagger}$ \quad \textbf{Mohak Shah}$^{\dagger*}$\\
$^{\dagger}$ Robert Bosch Research and Technology Center, CA \\ 
$^{\ddagger}$ University of Minnesota, MN \\
$^{*}$ University of Illinois at Chicago, IL \\ 
\texttt{\{sauptik.dhar, naveen.ramakrishnan, mohak.shah\}@us.bosch.com} \\ \texttt{cherk001@umn.edu}}

\begin{document}

\maketitle

\begin{abstract}
We introduce Universum learning \cite{vapnik98},\cite{vapnik06} for multiclass problems and propose a novel formulation for multiclass universum SVM (MU-SVM). We also propose a span bound for MU-SVM that can be used for model selection thereby avoiding resampling. Empirical results demonstrate the effectiveness of MU-SVM and the proposed bound.
\end{abstract}

\section{Introduction} \label{intro}
Many applications of machine learning involve analysis of sparse high-dimensional data, where the number of input features is larger than the number of data samples. Such high-dimensional data sets present new challenges for most learning problems. Recent studies have shown Universum learning to be particularly effective for such high-dimensional low sample size data settings~\cite{sinz08,chen09,dhar15,lu14,qi14,shen12,wang14,zhang08,xu15,xu16,zhu16a,zhu16b}. But, most such studies pertaining to classification problems are limited to binary (`two'- class) classification problems. On the other hand, many practical applications involve discrimination for more than two categories. Typical examples include, speech recognition, object recognition from images, prognostic health management etc~\cite{cherkassky07,hastie01}. In order to incorporate \textit{a priori} knowledge (in the form of Universum data) for such applications, there is a need to extend Universum learning for multiclass problems. 

In this paper we mainly focus on formulating the universum learning for multiclass SVM under balanced settings with equal misclassification costs. Support Vector Machines (SVM) have gained enormous popularity in machine learning, statistics and engineering over the last decades and are being used in many real-world applications. Researchers have proposed several methods to solve a multiclass SVM problem. Typically these methods follow two basic approaches (see~\cite{hsu02,wang14} for more details). The first approach follows an \textit{ensemble based setting}, where several binary classifiers are combined to construct the multiclass classifier viz., one-vs-one, one-vs-all, directed acyclic graph SVM \cite{platt99}. Previous works, such as~\cite{sinz07b,chen09} which follow the ensemble based setting, focus on the binary universum learning paradigm and only provide some hints for their extensions to the multiclass problems. An alternative to the ensemble based setting is the \textit{direct approach}, where the entire multiclass problem is solved through a single larger optimization formulation (see \cite{vapnik98,crammer02,weston98}). In this paper we develop and discuss MU-SVM, a \textit{direct approach} for universum learning following the Crammer \& Singer's (C\&S) multiclass SVM formulation~\cite{crammer02}. 

The main contributions of this work can be summarized as follows: 1). We propose a new (direct) formulation for universum learning for SVM under the multiclass setting, 2). we show that MU-SVM could be solved efficiently using any standard multiclass SVM solver and, 3). we derive a new leave-one-out bound for MU-SVM which provides a computationally efficient mechanism to perform model selection compared to the classical resampling based approach.


The paper is organized as follows. Section \ref{MSVM} describes the widely used multiclass SVM formulation in \cite{crammer02}. Section \ref{MUSVM} formalizes the notion of Universum learning for multiclass problems and introduces the new MU-SVM formulation (in section \ref{MUSVM1}). A discussion on the computational implementation of the MU-SVM is provided in section \ref{MUSVM2}. We derive a new \textit{leave-one-out} bound for the MU-SVM formulation in section \ref{MUSVM3} , and provide a simple two-step strategy for model selection. Section \ref{results} provides the empirical results in support of the proposed strategy. Finally, conclusions are presented in section \ref{conc}.

\section{Multiclass SVM} \label{MSVM}

\begin{wrapfigure}{r}{4.5cm}
\vspace{-20pt}
\includegraphics[width=4.5cm]{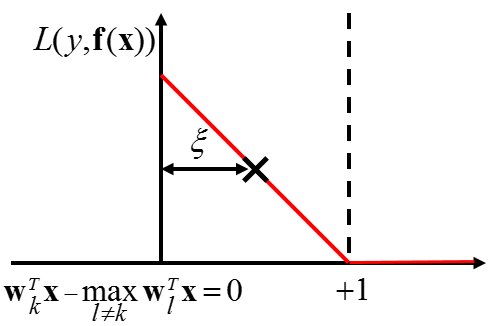}
\caption{Loss function for multiclass SVM with $f_k(\mathbf{x})=\mathbf{w}_k^{\top}\mathbf{x}$. A sample $(\mathbf{x},y=k)$ lying inside the margin is penalized linearly using the slack variable $\xi$.}\label{fig1}
\vspace{-22pt}
\end{wrapfigure} 

This section provides a brief description of the multiclass SVM formulation following Crammer \& Singer (C\&S) \cite{crammer02}. Given i.i.d training samples $(\mathbf{x}_i,y_i)_{i=1}^n$, with $\mathbf{x} \in \Re^d$ and $y \in \{1,\ldots, L\}$ ; where $n$ = number of training samples, $d$ = dimensionality of the input space and $L$ = total number of classes. The task of a multiclass classifier is to estimate a vector valued function $\mathbf{f} = [f_1,\ldots,f_L]$ for predicting the class labels for future unseen samples $(\mathbf{x},y)$ using the decision rule $\hat{y} = \underset{l=1,\ldots,L}{\text{argmax}}\; f_l(\mathbf{x})$. The C\&S multiclass SVM \cite{crammer02} is a widely used formulation which generalizes the concept of large margin classifier for  multiclass problems. This multiclass SVM setting employs a special margin-based loss (similar to the hinge loss), $L(y,\mathbf{f}(\mathbf{x})) = [\underset{l}{max}(f_l(\mathbf{x})+1-\delta_{yl})-f_y(\mathbf{x})]_{+}$ where $[a]_+ = max(0,a)$ and $\delta_{yl} = \left\{
\begin{array}{l l}
    1;\quad  y=l \\
    0;\quad  y\neq l
\end{array}\right.$ (see Fig \ref{fig1}). Here, for any sample $(\mathbf{x},y = k)$, having $L(y,\mathbf{f}(\mathbf{x})) = 0$ ensures a margin-distance of `+1' for the correct prediction i.e. $f_k(\mathbf{x})-f_l(\mathbf{x}) \geq 1 ; \forall l \neq k$. The SVM multiclass formulation (for linear parameterization) is provided below:
\begin{align}\label{eq1}
\underset{\mathbf{w}_1 \ldots \mathbf{w}_L ,\mathbf{\xi}}{\text{min}} & \quad \quad \frac{1}{2} \sum \limits_{l}\|\mathbf{w}_l \|_{2}^{2} \quad+\quad C\sum\limits_{i=1}^n \xi_{i} &&\\
s.t. & \quad \quad (\mathbf{w}_{y_{i}}-\mathbf{w}_l)^\top \mathbf{x}_i \geq e_{il} - \xi_{i} ;\quad e_{il} = 1-\delta_{il}; \quad i=1 \ldots n ,\quad l = 1 \ldots L  && \nonumber 
\end{align}
here,  $f_l(\mathbf{x}) = \mathbf{w}_l^{\top}\mathbf{x} $ and $\delta_{il} = \left\{
\begin{array}{l l}      
    1;\quad  y_i=l \\
    0;\quad  y_i\neq l
\end{array}\right.$ . Note that training samples falling inside the margin border (`+1') are linearly penalized using the slack variables $\xi_i \geq 0, i=1 \ldots n$ (as shown in Fig \ref{fig1}). These slack variables contribute to the empirical risk for the multiclass SVM formulation $R_{emp}(\mathbf{w})=\sum \limits_{i=1}^{n} \xi_i$. The SVM formulation attempts to strike a balance between minimization of the empirical risk and the regularization term. This is controlled through the user-defined parameter $C \geq 0$. For most SVM solvers eq. \eqref{eq1} is typically solved in it's dual form which provides a mechanism to extend the linear SVM to non-linear settings. This is accomplished by introducing a non-linear kernel function $K(\mathbf{x}_i,\mathbf{x}_j)=\langle \varphi(\mathbf{x}_i)\cdot\varphi(\mathbf{x}_j)\rangle$ that implicitly captures the non-linear mapping of the data $\mathbf{x}\rightarrow \varphi(\mathbf{x})$ (see \cite{crammer02} for more details). 

\section{Multiclass Universum SVM} \label{MUSVM}
\begin{wrapfigure}{r}{5cm}
\vspace{-70pt}
\includegraphics[width=5cm]{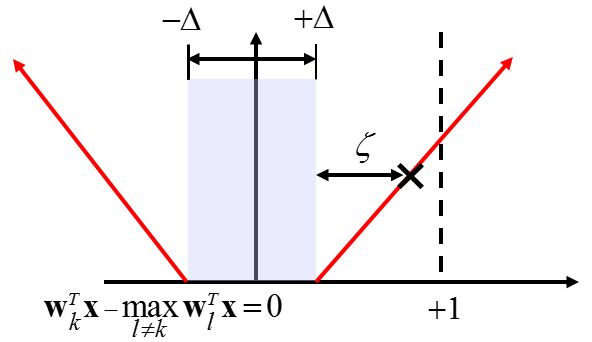}
\caption{Loss function for universum samples for $k^{th}$ decision function $f_k(\mathbf{x})=\mathbf{w}_k^{\top}\mathbf{x}$. A sample lying outside the $\Delta$- insensitive zone is penalized linearly using the slack variable $\zeta$.}\label{fig2}
\vspace{-10pt}
\end{wrapfigure}
\subsection{Multiclass U-SVM formulation} \label{MUSVM1}
The idea of Universum learning was introduced by Vapnik \cite{vapnik98,vapnik06} to incorporate a priori knowledge about admissible data samples. The Universum learning was introduced for binary classification, where in addition to labeled training data we are also given a set of unlabeled examples from the Universum. The Universum contains data that belongs to the same application domain as the training data. However, these samples are known not to belong to either class. In fact, this idea can also be extended to multiclass problems. For multiclass problems in addition to the labeled training data we are also given a set of unlabeled examples from the Universum. However, now the Universum samples are known not to belong to \textit{any} of the classes in the training data. For example, if the goal of learning is to discriminate between handwritten digits 0, 1, 2,...,9; one can introduce additional `knowledge' in the form of handwritten letters A, B, C, ... ,Z. These examples from the Universum contain certain information about handwriting styles, but they cannot be assigned to any of the classes (1 to 9). Also note that, Universum samples do not have the same distribution as labeled training samples. These unlabeled Universum samples are introduced into the learning as contradictions and hence should lie close to the decision boundaries for all the classes $\mathbf{f} = [f_1,\ldots,f_L]$. This argument follows from \cite{vapnik06,weston06}, where the universum samples lying close to the decision boundaries are more likely to falsify the classifier. To ensure this, we incorporate a $\Delta$ - insensitive loss function for the universum samples (shown in Fig \ref{fig2}). This $\Delta$ - insensitive loss forces the universum samples to lie close to the decision boundaries (`0' in Fig. \ref{fig2}). Note that, this idea of using a $\Delta$ - insensitive loss for Universum samples has been previously introduced in \cite{weston06} for binary classification. However, different from \cite{weston06}, here the $\Delta$ - insensitive loss is introduced for the decision functions of all the classes i.e.  $\mathbf{f}=[f_1,\ldots,f_L]$.  This reasoning motivates the new multiclass Universum-SVM (MU-SVM) formulation where:
\begin{itemize}
\item[--] Standard hinge loss is used for the training samples (shown in Fig. \ref{fig1}). This loss forces the training samples to lie outside the `+1' margin border.
\item[--] The universum samples are penalized by a $\Delta$ - insensitive loss (see Fig. \ref{fig2}) for the decision functions of all the classes $\mathbf{f} = [f_1,\ldots,f_L]$.
\end{itemize}
This leads to the following MU-SVM formulation. Given training samples $\mathcal{T}:=(\mathbf{x}_i,y_i)_{i=1}^n$, where $y_i \in \lbrace 1,\ldots,L\rbrace$ and additional unlabeled universum samples $\mathcal{U}:=(\mathbf{x}_j^{*})_{j=1}^m$. Solve~\footnote{Throughout this paper, we use index $i$ for training samples, $j$ for universum samples and $k, l$ for the class labels.},
\begin{align}\label{eq2}
\underset{\mathbf{w}_1 \ldots \mathbf{w}_L ,\mathbf{\xi},\mathbf{\zeta}}{\text{min}} & \quad \quad \frac{1}{2} \sum \limits_{l}\|\mathbf{w}_l \|_{2}^{2} \quad+\quad C\sum\limits_{i=1}^n \xi_{i} \quad+\quad C^{*}\sum\limits_{j=1}^m \zeta_{j}&&\\
s.t. & \quad \quad (\mathbf{w}_{y_{i}}-\mathbf{w}_l)^\top \mathbf{x}_i \geq e_{il} - \xi_{i} ;\quad e_{il} = 1-\delta_{il}, \quad i=1 \ldots n  && \nonumber  \\
& \quad \quad |(\mathbf{w}_k-\mathbf{w}_l)^\top \mathbf{x}_j^{*}| \leq \Delta + \zeta_{j}; \quad j=1 \ldots m, \quad l,k = 1 \ldots L  && \nonumber 
\end{align}
Here, the universum samples that lie outside the $\Delta$ - insensitive zone are linearly penalized using the slack variables $\zeta_j \geq 0, j = 1\ldots m$. The user-defined parameters $C,C^{*} \geq 0$ control the trade-off between the margin size, the error on training samples, and the contradictions (samples lying outside $\pm\Delta$ zone) on the universum samples. Note that for $C^{*} = 0$ this formulation becomes equivalent to the multiclass SVM classifier.

\subsection{Computational Implementation of MU-SVM} \label{MUSVM2}

In this section we discuss the current implementation of the MU-SVM formulation in \eqref{eq2}. Following \cite{weston06}, for each universum sample $(\mathbf{x}^{*})$  we create artificial samples belonging to all the classes, i.e. $(\mathbf{x}_j^{*},y_j^{*}=1),\ldots, (\mathbf{x}_j^{*},y_j^{*}=L)$. For simplicity we overload the variables as shown below:
\begin{flalign} \label{eq3}
\mathbf{x}_i &= \left\{
\begin{array}{l l l} 
\mathbf{x}_i \quad & i=1 \ldots n \quad \text{(training samples)} \\
\mathbf{x}_j^{*} \quad & i=n+1 \ldots n+mL;\ j=1 \ldots mL \quad \text{(universum samples)}
\end{array} \right. && \nonumber \\
y_i &= \left\{
\begin{array}{l l l} 
y_i \quad & i=1 \ldots n  \\
y_j^{*}   \quad & i=n+1 \ldots n+mL;\ j=1 \ldots mL
\end{array} \right. && \nonumber \\
e_{il} &= \left\{
\begin{array}{l l l} 
e_{il} \quad & i=1 \ldots n; \quad l=1 \ldots L  \\
-\Delta(1-\delta_{jl}) \quad & i=n+1 \ldots n+mL;\quad j=1 \ldots mL; \quad l=1 \ldots L
\end{array} \right. &&  \\
C_i &= \left\{
\begin{array}{l l l} 
C \quad & i=1 \ldots n \\
C^{*} \quad & i=n+1 \ldots n+mL; \quad j=1 \ldots mL
\end{array} \right. && \nonumber \\
\xi_i &= \left\{
\begin{array}{l l l} 
\xi_i \quad & i=1 \ldots n \\
\zeta_j \quad & i=n+1 \ldots n+mL; \quad j=1 \ldots mL
\end{array} \right. && \nonumber 
\end{flalign}   
Then \eqref{eq2} can be re-written as,
\begin{align}\label{eq4}
\underset{\mathbf{w}_1 \ldots \mathbf{w}_L ,\mathbf{\xi}}{\text{min}} & \quad \quad \frac{1}{2} \sum \limits_{l}\|\mathbf{w}_l \|_{2}^{2} \quad+\quad \sum\limits_{i=1}^{n+mL} C_i \ \xi_{i} &&\\
s.t. & \quad \quad (\mathbf{w}_{y_{i}}-\mathbf{w}_l)^\top \mathbf{x}_i \geq e_{il} - \xi_{i}  \quad i=1 \ldots n+mL ,\quad l = 1 \ldots L && \nonumber
\end{align}
The formulation~\eqref{eq4} has the same form as~\eqref{eq1} except that the former has additional $mL$ constraints for the universum samples. Like most other SVM solvers, the MU-SVM formulation in~\eqref{eq4} is also solved in its dual form (shown in Algorithm~\ref{alg1}). Hence, the computational complexity is same as solving a multiclass SVM formulation (in~\eqref{eq1}) with $n+mL$ samples. Most off-the-shelf multiclass SVM solvers can be used for solving the proposed MU-SVM. For completeness, we show the steps for the proposed MU-SVM solver in Algorithm~\ref{alg1}:

\begin{algorithm} \SetAlgoNoLine 
1. Given training $(\mathbf{x}_i,y_i)_{i=1}^n$ and universum samples $(\mathbf{x}_j^{*})_{j=1}^m$ perform the transformation in \eqref{eq3} \;
2. Solve \eqref{eq5} to obtain the MU-SVM solution, \
\begin{align}\label{eq5}
\underset{\boldsymbol\alpha}{\text{max}} & \quad \quad W(\boldsymbol \alpha)= - \frac{1}{2} \sum \limits_{i,j} \sum \limits_{l} \alpha_{il} \alpha_{jl} K(\mathbf{x}_i,\mathbf{x}_j)\quad - \quad \sum\limits_{i,l}\alpha_{il}e_{il} &&\\
s.t. & \quad \sum \limits_{l} \alpha_{il} =0  ; \quad \alpha_{i,l} \leq C_i \quad  \text{if} \quad l=y_i \quad ; \quad \alpha_{i,l} \leq 0 \quad  \text{if} \quad l \neq y_i  && \nonumber 
\end{align} \
3. Obtain the class label using the following decision rule:\quad $\hat{y} = \underset{l}{\text{argmax}}\sum \limits_{i}\alpha_{il}K(\mathbf{x}_i,\mathbf{x})$\
\caption{MU-SVM (dual form) \label{alg1}}
\end{algorithm}

\subsection{Model Selection} \label{MUSVM3}
As presented in~\eqref{eq5}, the current MU-SVM algorithm has four tunable parameters: $C, C^* ,\text{kernel parameter, and }  \Delta$. So in practice, multiclass SVM may yield better results than MU-SVM, simply because it has an inherently simpler model selection. A successful practical application of the proposed MU-SVM heavily depends on the optimal tuning of the model parameters. This paper proposes to adopt a simplified strategy (previously used in \cite{cherkassky11,dhar15}) for model selection which mainly involves two steps,
\begin{itemize}
\item[Step a.]  First, perform optimal tuning of the $C$ and kernel parameters for multiclass SVM classifier. This step equivalently performs model selection for the parameters specific only to the training samples in the MU-SVM formulation \eqref{eq2}.
\item[Step b.] Second, tune the parameter $\Delta$ while keeping $C$ and kernel parameters fixed (as selected in Step a). Parameter $C^*/C = \frac{n}{mL}$ is kept fixed throughout this paper to have equal weightage on the loss due to training and universum samples.
\end{itemize}
The model parameters in Steps (a) \& (b) are typically selected through resampling techniques or using a separate validation set. In this paper however, we provide a new analytic bound for the leave-one-out error (l.o.o) for MU-SVM formulation. Note that, by removing the universum samples, we obtain the l.o.o bound for the multiclass SVM formulation. Now, the model parameters in Steps (a) \& (b) are selected to minimize this leave-one-out (l.o.o) error bound. A detailed discussion regarding this new l.o.o error bound is provided next.

Note that, the l.o.o formulation with the $t^{th}$ training sample dropped is the same as in \eqref{eq5} with an additional constraint $\alpha_{tl} = 0 ; \quad  \forall l$. Then, the l.o.o error is given as: $ R_{l.o.o} = \frac{1}{n}\sum \limits_{t=1}^{n} \mathbbm{1}[y_t \neq \hat{y}_t] $, where $\hat{y}_t=  \underset{l}{\text{arg max}}\sum \limits_{i}\alpha_{il}^t K(\mathbf{x}_i,\mathbf{x}_t)$ is the predicted class label for the $t^{th}$ sample and $\boldsymbol\alpha^t = [\underset{\boldsymbol\alpha_1^t}{\underbrace{\alpha_{11}^t,\ldots,\alpha_{1L}^t}},\ldots,\underset{\boldsymbol\alpha_t^t = \mathbf{0}}{\underbrace{\alpha_{t1}^t = 0,\ldots,\alpha_{tL}^t = 0}},\ldots]$ is the l.o.o solution. In this paper we follow a very similar strategy as used in \cite{vapnik00}, and derive the new l.o.o bound for the MU-SVM formulation in \eqref{eq5}. The necessary prerequisites are presented next.
\begin{definition}(Support vector categories)\label{def1} 
\begin{itemize}
\item[Type 1.] A support vector obtained from eq.~\eqref{eq5} is called a \textit{Type 1} support vector if $0 < \alpha_{iy_i} < C_i$. This is represented as, $SV_1 = \{\ i \ | 0< \alpha_{iy_i} < C_i \}$ 
\item[Type 2.] A support vector obtained from eq.~\eqref{eq5} is called a \textit{Type 2} support vector if $
\alpha_{iy_i}=C_i$. This is represented as, $SV_2 = \{\ i \ | \alpha_{iy_i} = C_i \}$ 
\end{itemize}
\end{definition}
The set of all support vectors are represented as, $SV = SV_1 \cup SV_2$. Similarly, the set of support vectors for \textit{l.o.o} solution is given as $SV^t$. Under \textit{definition}~\eqref{def1} we make the following assumptions.
\newline
\begin{assumption} \label{as1}
The set of support vectors of the \textit{Type1} and \textit{Type2} categories remain the same during the leave-one-out procedure.
\end{assumption}

This is a well-established assumption which has been previously used to derive the \textit{l.o.o} bound for binary SVM in~\cite{vapnik00}. The advantage of this assumption is that it reduces the computational complexity of the \textit{l.o.o} bound (see Corollary~\eqref{col2}). However, in this paper we make an additional assumption as given below.
\newline
\begin{assumption} \label{as2}
The dual variables of the \textit{Type1} support vectors have only two active elements i.e. $\forall \boldsymbol\alpha_i ~\textit{s.t.}~ \lbrace 0<\alpha_{iy_i}<C_i \rbrace \ \exists \ k \neq y_i ~~\textit{s.t.} ~~ \alpha_{ik} = - \alpha_{iy_i}$.
\end{assumption}
This assumption provides the advantage of analyzing the bound for the multiclass problem in a one-vs-one (binary) fashion. We observe that, under high-dimensional low sample size settings this assumption holds true for almost all Type 1 support vectors. A more detailed analysis shall be provided in a longer version of this paper. Next we provide the main result used for the leave-one-out error bound.
\newline
\begin{theorem} \label{th1}
Under Assumptions 1\& 2 the following equality holds for the Type 1 support vectors $\forall \boldsymbol\alpha_t ~\textit{s.t.}~ \lbrace 0<\alpha_{ty_t}<C_t \rbrace $,
\begin{align}\label{eq6}
S_t^2 = [\boldsymbol\alpha_t^{\top} \sum\limits_{i \in SV} \sum\limits_{l} \alpha_{il} K(\mathbf{x}_i,\mathbf{x}_t) - \alpha_{ty_t}\mathbf{g}_k^{\top} \sum \limits_{i \in SV^t} \sum \limits_{l} \alpha_{il}^t K(\mathbf{x}_i,\mathbf{x}_t)]
\end{align}
\end{theorem} where, \quad  $S_t^2 =\{\underset{\boldsymbol\beta}{\text{min}}\ \sum \limits_{i,j}(\sum \limits_{l}\beta_{il}\beta_{jl})K(\mathbf{x}_i,\mathbf{x}_j)|\ \boldsymbol\beta_{t} =\boldsymbol\alpha_{t} ;\ \sum \limits_{l} \beta_{il} =0\ ; (i,j)\in SV_1\}$\quad and \quad  $\mathbf{g}_k =[0,\ldots \underset{l^{th}=y_t}{1},\ldots,\underset{k^{th}}{-1},\ldots,0]$. \\
\textbf{Proof} See supplementary material. \\
Note that, this equality is very similar to the result in \cite{vapnik00} obtained for binary SVM. Same as \cite{vapnik00} we refer to $S_t$ as the (constrained) span of the \textit{Type 1} support vectors. However, for practical cases the computation of $S_t$ can be simplified following the corollary \eqref{col1}. 
\newline
\begin{corollary} \label{col1}
The span $S_t^2$ can be efficiently computed as
\begin{align} \label{eq7}
S_t^2 &= \boldsymbol\alpha_t^{\top} [(\mathbf{H}^{-1})_{\mathbf{tt}}]^{-1} \boldsymbol\alpha_t
\end{align}
\begin{flalign}
\text{here,} \quad \mathbf{H} & := \begin{bmatrix}
    \mathbf{K}_{SV_1} \otimes \mathbf{I}_{L}       & \mathbf{A}^{\top} \\
    \mathbf{A}     & \mathbf{0}
\end{bmatrix}; \quad \quad \mathbf{A} := \mathbf{I}_{|SV_1|} \otimes (\mathbf{1}_L)^{\top} ; \quad \quad  \mathbf{1}_L =[\underset{L\ elements}{\underbrace{1 \ 1 \ldots \ 1}}] \nonumber && \\
\quad (\mathbf{H}^{-1})_{\mathbf{tt}} \quad & := \text{sub-matrix of } \;  \mathbf{H}^{-1} \ \text{for index }\;   i\  = (t-1)L+1,\ldots ,tL &&\nonumber \\
\quad \mathbf{K}_{SV_1} \quad & := \text{Kernel matrix of Type 1 support vectors}. &&\nonumber 
\end{flalign}
\end{corollary}  
where, $\otimes$ is the Kronecker product. \newline
\textbf{Proof} See supplementary material. \newline
Now rather than solving a quadratic program (as shown in Theorem \eqref{th1}), the computation of $S_t$ mainly involves computing the inverse of the $\mathbf{H}$ - matrix. This is an $O(n+mL)^3$ operation, and provides a computational advantage over computing the \textit{l.o.o} error, which is $O(n+mL)^4$. Finally, we use the results in Theorem \eqref{th1} and Corollary \eqref{col1} to obtain the following,
\newline
\begin{corollary} \label{col2}
Under the Assumptions 1 \& 2 the leave-one-out error is upper bounded as: 
\begin{flalign} \label{eq8}
R_{l.o.o} & \leq \frac{1}{n} [ \; \; Card \{ \; t \; | \; \boldsymbol\alpha_t^{\top} [(\mathbf{H}^{-1})_{\mathbf{tt}}]^{-1} \boldsymbol\alpha_t \geq  \boldsymbol\alpha_t^{\top} \sum\limits_{i \in SV} \sum\limits_{l} \alpha_{il} K(\mathbf{x}_i,\mathbf{x}_t)\; ; \;t \in SV_1 \cap \mathcal{T} \} &&\\
& \quad  \quad +  Card \{ \; t \; | \;  \;t \in  SV_2 \cap \mathcal{T} \} ] ; \quad \quad \text{where} \; \mathcal{T}:=\text{Training Set} && \nonumber 
\end{flalign}
\end{corollary}  \textbf{Proof} See supplementary material. \newline
For the rest of the paper we use the eq. \eqref{eq8} for model selection in Steps a \& b and select the parameters which minimizes the right hand side of the bound in eq. \eqref{eq8}.

\section{Empirical Results} \label{results}
\begin{wrapfigure}{r}{3cm}
\vspace{-5pt}
\begin{tabular}{@{}c@{}}
    \includegraphics[width=3cm]{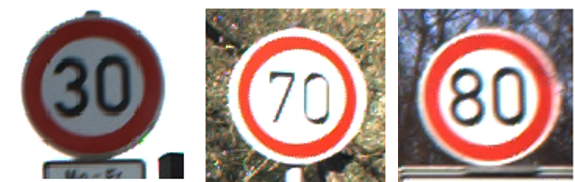} \\
    \text{(a) Training samples} \\
    \includegraphics[width=2cm]{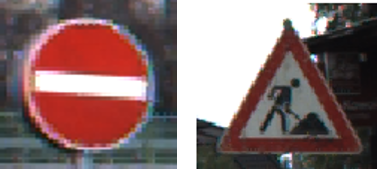} \\
    \text{(b) Universum samples} \\
\end{tabular}
\caption{GTSRB data.}\label{fig3}
\vspace{-15pt}
\end{wrapfigure}

Our empirical results mainly use two real life datasets: 
\\[0.4em]
\textit{German Traffic Sign Recognition Benchmark (GTSRB) dataset} \cite{stall12} : The goal here is to identify the traffic signs `30',`70' and `80' (shown in Fig.\ref{fig3}a). Here, the sample images are represented by their histogram of gradient (HOG) features (following \cite{shen12,dhar15}). Further, in addition to the training samples we are also provided with additional universum samples i.e. traffic signs for \textit{no-entry}' and `\textit{roadworks}'(shown in Fig.\ref{fig3}b). Note that these universum samples belong to the same application domain i.e. they are traffic sign images. However, they do not belong to any of the training classes. Analysis using the other types of Universum have been omitted due to space constraints.
\begin{wrapfigure}{r}{3cm}
\vspace{-25pt}
\begin{tabular}{@{}c@{}}
    \includegraphics[width=3cm]{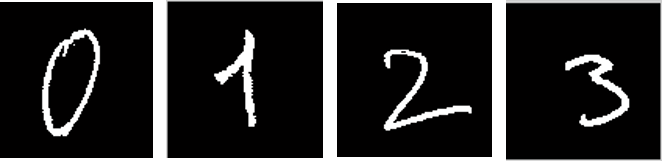} \\
    \text{(a) Training samples} \\
    \includegraphics[width=1.5cm]{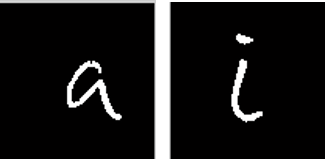} \\
    \text{(b) Universum samples} \\
\end{tabular}
\caption{ABCDETC dataset.}\label{fig4}
\vspace{-35pt}
\end{wrapfigure}

\textit{Real-life ABCDETC dataset} \cite{weston06}: This is a handwritten digit recognition dataset, where in addition to the digits `0-9' we are also provided with the images of the \textit{uppercase}, \textit{lowercase} handwritten letters and some additional special symbols. In this paper, the goal is to identify the handwritten digits `0' - `3' based on their pixel values. Further, we use the images of the handwritten `\textit{letters a}' and `\textit{i}' as universum samples for illustration.

The experimental settings used for these datasets throughout the paper is provided in Table \ref{tab1}. For the GTSRB dataset we have performed number of experiments with varying universum set sizes and provide the optimal set size in Table \ref{tab1}. Further increase in the number of universum samples did not provide significant performance gains (see supplementary material for more details). 

\begin{table}
\centering
\caption{Experimental settings for the Real-life datasets.} \label{tab1}
\begin{tabular}{|c||c|c|c|c|} \hline 
  \textbf{Dataset} & \textbf{Training size} & \textbf{Test size} & \textbf{Universum size} & \textbf{Dimension}\\ \hline
  GTSRB & \specialcell{300\\(100 per class)} & \specialcell{1500\\(500 per class)}& 500 &\specialcell{1568\\(HOG Features)}\\ \hline
  ABCDETC & \specialcell{600\\(150 per class)} & \specialcell{400\\(100 per class)}& 250\textsuperscript{*} & \specialcell{10000\\(100 x 100 pixel)}\\ \hline
\multicolumn{5}{l}{\textsuperscript{*}\footnotesize{used all available samples.}}
\end{tabular}
\end{table}

\subsection{Comparison between Multiclass SVM vs. U-SVM}
Our first set of experiment uses the GTSRB dataset. Initial experiments suggest that linear parameterization is optimal for this dataset; hence only linear kernel has been used. Here, the model selection is done over the range of parameters, $C = [10^{-4},\ldots,10^{3}]$ , $C^{*}/C = \frac{n}{mL} = 0.2 $ and $\Delta =[0,0.01,0.05,0.1]$ using stratified 5-Fold cross validation \cite{shah11}. Performance comparisons between SVM and U-SVM for the different types of Universum: signs `\textit{no-entry}', and `\textit{roadworks}' are shown in Table \ref{tab2}.  The table shows the average $ \text{Test Error} = \frac{1}{n_T}\sum \limits_{i=1}^{n_T} \mathbbm{1}[y_i^{test} \neq \hat{y}_i^{test}] $ over 10 random training/test partitioning of the data in similar proportions as shown in Table. \ref{tab1}. Here $y_i^{test}\sim$ class label for $i^{th}$ test sample, $\hat{y}_i^{test}\sim$ predicted label for $i^{th}$ test sample and $n_T = $ number of test samples.

\begin{figure}
\centering
\includegraphics[height=2cm, width=10cm]{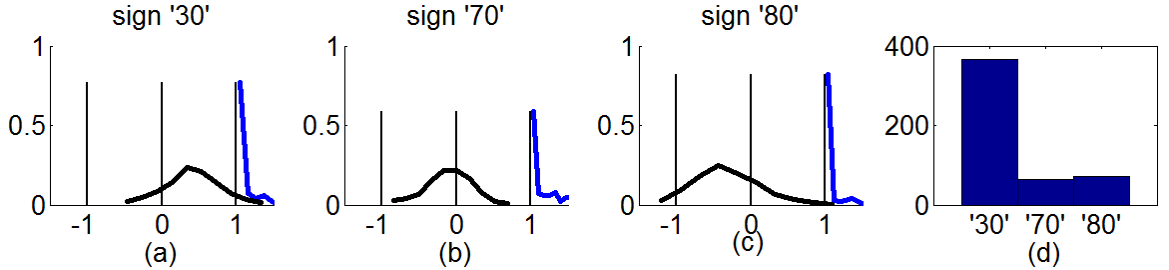}
\caption{Typical histogram of projection of training samples (shown in \textcolor{blue}{\textbf{blue}}) and universum samples (shown in \textbf{black}) onto the multiclass SVM model (with $C=1$). Decision functions for (a) sign `30'. (b) sign `70'.(c) sign `80'. (e) frequency plot of predicted labels for universum samples.} \label{histSVM1}
\includegraphics[height=2cm, width=10cm]{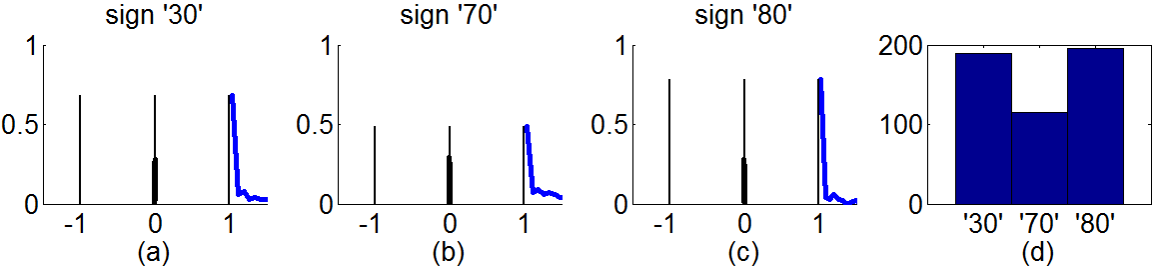}
\caption{Typical histogram of projection of training samples (shown in \textcolor{blue}{\textbf{blue}}) and universum samples (shown in \textbf{black}) onto the  MU-SVM model (with $\Delta=0$). Decision functions for (a) sign `30'. (b) sign `70'.(c) sign `80'. (e) frequency plot of predicted labels for universum samples.} \label{histUSVM1}
\end{figure}

As seen from Table \ref{tab2}, the MU-SVM models using both types of Universa provides better generalization than the multiclass SVM model. Here, for all the methods we have training error $\sim$ 0\%. For better understanding of the MU-SVM modeling results we adopt the technique of `\textit{histogram of projections}' originally introduced for binary classification \cite{cherkassky11,cherk10}. However, different from binary classification, here we project a training sample $(\mathbf{x},y = k)$ onto the decision space for that class i.e. $\mathbf{w}_k^{\top}\mathbf{x}-\underset{l \neq k}{\text{max}}\ \mathbf{w}_l^{\top}\mathbf{x}=0$ and the universum samples onto the decision spaces of all the classes. Finally, we generate the histograms of the projection values for our analysis. In addition to the histograms, we also generate the frequency plot of the predicted labels for the universum samples. Figs \ref{histSVM1} and \ref{histUSVM1} shows the typical histograms and frequency plots for the SVM and MU-SVM models using the `\textit{no-entry}' sign (as universum). As seen from Fig. \ref{histSVM1}, the optimal SVM model has high separability for the training samples i.e., most of the training samples lie outside the margin borders with training error $\sim$ 0. Infact, similar to binary SVM \cite{cherk10}, we see data-piling effects for the training samples near the `+1' - margin borders of the decision functions for all the classes. This is typically seen under high-dimensional low sample size settings. However, the universum samples (sign `\textit{no-entry}') are widely spread about the margin-borders. Moreover, for this case the universum samples are biased towards the positive side of the decision boundary of the sign `30' (see Fig \ref{histSVM1}(a)) and hence predominantly gets classified as sign `30'(see Fig.\ref{histSVM1} (d)). As seen from Figs  \ref{histUSVM1} (a)-(c), applying the MU-SVM model preserves the separability of the training samples and additionally reduces the spread of the universum samples. For such a model the uncertainity due to universum samples is uniform across all the classes i.e. signs `30',`70' and `80' (see Fig. \ref{histUSVM1}(d)). The resulting MU-SVM model has higher contradiction on the universum samples and provides better generalization in comparison to SVM. The histograms for the multiclass SVM and MU-SVM models using the sign `\textit{roadworks}' as universa are provided in supplementary material.

\begin{figure}
\centering
\includegraphics[height=1.5cm, width=12cm]{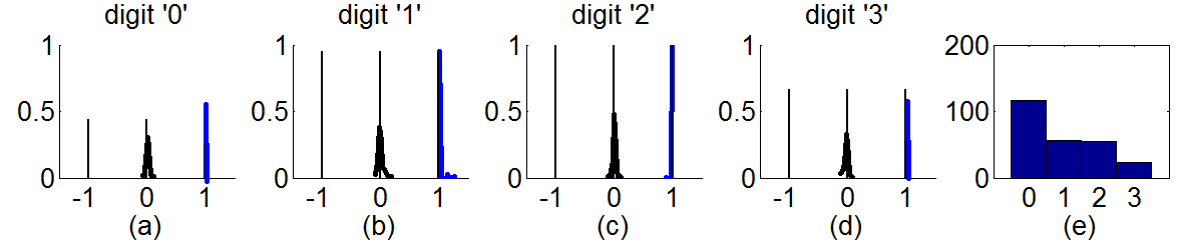}
\caption{Typical histogram of projection of training samples (in \textcolor{blue}{\textbf{blue}}) and universum samples (in \textbf{black}) onto the SVM model (with $C=1$ and $\gamma = 2^{-7} $). (a) digit `0'. (b) digit `1'.(c) digit `2'. (d) digit `3'. (e) frequency plot of predicted labels for universum samples (lowercase letter `a').} \label{histSVMa}
\includegraphics[height=1.5cm, width=12cm]{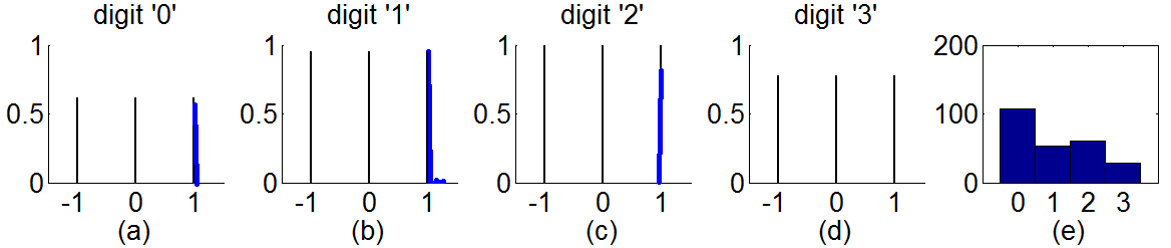}
\caption{Typical histogram of projection of training samples (in \textcolor{blue}{\textbf{blue}}) and universum samples (in \textbf{black}) onto MU-SVM model (with $C^{*}/C=0.6$ and $\Delta = 0.1$ ). (a) digit `0'. (b) digit `1'.(c) digit `2'. (d) digit `3'.(e) frequency plot of predicted labels for universum samples (lowercase letter `a').} \label{histUSVMa}
\end{figure}

Our \textit{next} experiment uses the ABCDETC dataset. For this dataset, using an RBF kernel of the form $K(\mathbf{x}_i,\mathbf{x}_j) = exp(-\gamma \Vert \mathbf{x}_i - \mathbf{x}_j \Vert^2)$ with $\gamma = 2^{-7}$ provided optimal results for SVM. The model selection is done over the range of parameters, $C = [10^{-4},\ldots,10^{3}]$, $C^*/C = 0.6 $ and $\Delta =[0,0.01,0.05,0.1]$ using stratified 5-Fold cross validation. Performance comparisons between multiclass  SVM and MU-SVM for the different types of Universum: letters `\textit{a}', and `\textit{i}' are shown in Table \ref{tab2}. In this case, MU-SVM using letter `\textit{i}' provides an improvement over the multiclass SVM solution. However, using letter `\textit{a}' as universum does not provide any improvement over the SVM solution. For better understanding we analyze the histogram of projections and the frequency plots for the multiclass SVM/MU-SVM models using the letter `\textit{a}' as universum in Figs. \ref{histSVMa},\ref{histUSVMa}. As seen in Fig. \ref{histSVMa} (a)-(d)) the SVM model already results in a narrow distribution of the universum samples and in turn provides \textit{near} random prediction on the universum samples (Fig. \ref{histSVMa}(e)). Applying MU-SVM for this case provides no significant change to multiclass SVM solution and hence no additional improvement in generalization (see Table \ref{tab2} and Fig. \ref{histUSVMa}). Finally, the histograms for the multiclass SVM/MU-SVM models using \textit{letters `i'} as universum display similar properties as in Figs \ref{histSVM1} \& \ref{histUSVM1} (please refer to supplementary material).


\begin{table} 
\centering
\caption{Performance comparisons between multiclass SVM vs. MU-SVM. The results show mean test error in \%, over 10 runs. The numbers in parentheses denote the standard deviations.} \label{tab2}
\begin{tabular}{|c||c|c|c|} 
\hline 
\textbf{Dataset} & \textbf{SVM}  & \textbf{MU-SVM} & \textbf{MU-SVM}  \\
\hline
GTSRB & 7.47 (0.92) & (sign `\textit{no-entry}'):  \textbf{6.57} (\textbf{0.59}) & (sign `\textit{roadworks}'): 6.88 (0.87)\\ 
\hline
ABCDETC &  26.15 (2.08) &(letter `\textit{a}'): 25.35 (2.13) &(letter `\textit{i}'): \textbf{22.05} (\textbf{2.07})\\
\hline
\end{tabular}
\end{table} 

The results in this section shows that MU-SVM provides better performance than multiclass SVM, typically for high-dimensional low sample size settings. Under such settings the training data exhibits large data-piling effects near the margin border (`+1'). For such ill-posed settings, introducing the Universum can provide improved generalization over the multiclass SVM solution. However, the effectiveness of the MU-SVM also depends on the properties of the universum data. Such statistical characteristics of the training and universum samples for the effectiveness of MU-SVM can be conveniently captured using the `histogram-of-projections' method introduced in this paper.

\subsection{Effectiveness of the Model Selection using Bound in \eqref{eq8}}
Next we provide results showing the practical utility of the bound in \eqref{eq8} for model selection. Here, we provide the performance results of the MU-SVM model when the model parameters are selected using \eqref{eq8}. That is, we select the model parameters which provides the smallest value for the bound \eqref{eq8}. Table \ref{tab3} shows the average test error over 10 random training/test partitioning of the data in similar proportions as shown in Table. \ref{tab1}. As seen from Table \ref{tab3}, the MU-SVM models selected using \eqref{eq8} provides comparable results to the standard 5-Fold resampling technique. This provides a practical alternative for model selection for the MU-SVM algorithm. \footnote{Modeling results using \textit{l.o.o} strategy was prohibitively slow, and hence could not be reported in this paper.} The proposed model selection strategy using \eqref{eq8} involves an $O(n+mL)^3$ operation, and provides a computational edge over standard resampling techniques. For example, the average time complexity over 10 experiments (GTSRB with Universum:'no-entry') shows, MUSVM(CV) $\sim$ 4413s vs. MUSVM(bound) $\sim$ 1583s. Detailed time complexity analysis and results on the LOO bound shall be provided in a longer version of this paper.
\begin{table} 
\centering
\caption{Performance results for MU-SVM using the bound in eq. \eqref{eq8}. The results show mean test error in \%, over 10 runs. The numbers in parentheses denote the standard deviations. Learning by bound minimization performs as good as, or better than resampling based model selection.} \label{tab3}
\begin{tabular}{|c||*{2}{c|}}\hline
\textbf{Dataset (Universum)}
& \specialcell{\textbf{MU-SVM (Resampling)}\\(Test Error in \%)} & \specialcell{\textbf{MU-SVM (Bound)}\\(Test Error in \%)}\\\hline
GTSRB (`\textit{no-entry}') & 6.57 (0.59) & 6.47 (0.57)\\\hline
GTSRB (`\textit{roadworks}')& 6.88 (0.87) & 6.9 (0.75)\\\hline
ABCDETC (`letter \textit{a}') &25.35 (2.13)& 25.25 (2.03)\\\hline
ABCDETC (`letter \textit{i}') &22.05 (2.07)& 22.40 (1.73)\\\hline
\end{tabular}
\end{table}


\section{Conclusions} \label{conc}
We introduced universum learning for multiclass problems and provided a new universum-based formulation for multiclass SVM (MU-SVM). This formulation reduces to the classical multiclass SVM formulation in the absence of universum samples and can utilize standard SVM solvers. We also proposed a novel span bound for the MU-SVM that can be used to perform efficient model selection. We empirically demonstrated the effectiveness of the proposed formulation as well as the bound on real-world datasets. In addition, we also provided insights into the underlying behavior of universum learning and its dependence on the choice of universum samples using the proposed `histogram-of-projections' method. 

\small
\bibliographystyle{IEEEtran}  
\bibliography{MultiClassUSVM}

\appendix
\input{MultiClassUSVM_appendix.tex}
\end{document}

%% file: MultiClassUSVM_appendix.tex
\section{Proofs}
\textit{The references cited in this document follows the numbering used in the main paper}.

\subsection{Proof of Theorem 1}
The proof follows similar lines as in [21]. As previously discussed in Section \ref{MUSVM3}, the \textit{leave-one-out} formulation for U-SVM with the $t^{th}$ sample dropped is,
\begin{align}\label{eq9}
\underset{\boldsymbol\alpha}{\text{max}} & \quad \quad W(\boldsymbol\alpha)= - \frac{1}{2} \sum \limits_{i,j} \sum \limits_{l} \alpha_{il} \alpha_{jl} K(\mathbf{x}_i,\mathbf{x}_j)\quad - \quad \sum\limits_{i,l}\alpha_{il}e_{il} &&\\
s.t. & \quad \sum \limits_{l} \alpha_{il} =0 ;\quad \alpha_{il} \leq C_i \quad  \text{if} \quad l=y_i \quad ; \quad \alpha_{il} \leq 0 \quad  \text{if} \quad l \neq y_i  && \nonumber \\
& \quad \alpha_{tl} = 0 ; \quad  \forall l  \quad \text{(additional constraint)}&& \nonumber
\end{align} 
Then, the \textit{leave-one-out} (l.o.o) error is given as: $ R_{l.o.o} = \frac{1}{n}\sum \limits_{t=1}^{n} \mathbbm{1}[y_t \neq \hat{y}_t] $ where, $\boldsymbol\alpha^t = [\underset{\boldsymbol\alpha_1^t}{\underbrace{\alpha_{11}^t,\ldots,\alpha_{1L}^t}},\ldots,\underset{\boldsymbol\alpha_t^t = \mathbf{0}}{\underbrace{\alpha_{t1}^t = 0,\ldots,\alpha_{tL}^t = 0}},\ldots]$ is the solution for \eqref{eq9} and $\hat{y}_t=  \underset{l}{\text{arg max}}\sum \limits_{i}\alpha_{il}^t K(\mathbf{x}_i,\mathbf{x}_t)$ (estimated class label for the $t^{th}$ sample). The overall proof for the bound on the l.o.o error follows three major steps. 

\textit{First}, we construct a feasible solution for \eqref{eq5} using the optimal leave-one-out solution $\boldsymbol\alpha^t$. i.e., construct $\boldsymbol\alpha^t + \boldsymbol\gamma$ as shown below,
\begin{flalign}\label{eq10}
\quad \alpha_{il}^t + \gamma_{il} \leq C_i; \quad & \quad \forall\ (i,l) \in \lbrace (i,l) |\ \alpha_{il}^t < C_i;\ l = y_i \rbrace := A_1^t && \nonumber \\
\quad \alpha_{il}^t + \gamma_{il} \leq 0 ;\quad & \quad \forall\ (i,l) \in \lbrace (i,l) |\ \alpha_{il}^t < 0;\ l \neq y_i \rbrace := A_2^t && \nonumber \\
\quad \sum \limits_{l} \gamma_{il} =0 ;\quad \quad &&& \nonumber
\end{flalign}
and,
\begin{flalign}
\quad \gamma_{il} =0  \quad  \quad & \quad \forall (i,l) \notin SV_1^t &[\text{with}\ SV_1^t = A_1^t \cup A_2^t = \{\ i \ | 0< \alpha_{iy_i}^t < C_i \}] && \\
\quad \gamma_{tl} =\alpha_{tl} \quad & \quad \forall l & (under Assumption 2)&& \nonumber
\end{flalign}
such that, it is a feasible solution for \eqref{eq5}. Now,
\begin{flalign} \label{eq11}
I_1 &= W(\boldsymbol\alpha^t + \boldsymbol\gamma)-W(\boldsymbol\alpha^t) && \nonumber\\
&=-\frac{1}{2}\sum \limits_{i,j} \sum \limits_{l}(\alpha_{il}^t + \gamma_{il})(\alpha_{jl}^t + \gamma_{jl})K(\mathbf{x}_i,\mathbf{x}_j)-\sum \limits_{i} \sum \limits_l (\alpha_{il}^t + \gamma_{il})e_{il} &&\nonumber\\
&\quad +\frac{1}{2} \sum \limits_{i,j} \sum \limits_l \alpha_{il}^t \alpha_{jl}^t K(\mathbf{x}_i,\mathbf{x}_j) + \sum \limits_i \sum \limits_l \alpha_{il}^t e_{il}&& \nonumber \\ 
&= -\frac{1}{2}\sum \limits_{i,j} (\sum \limits_{l} \gamma_{il} \gamma_{jl}) K(\mathbf{x}_i,\mathbf{x}_j)-  \sum \limits_{i,j} (\sum \limits_{l}\gamma_{il} \alpha_{jl}^t)K(\mathbf{x}_i,\mathbf{x}_j)-\sum \limits_{i}\sum \limits_{l}\gamma_{il}e_{il} && \nonumber \\
&= -\frac{1}{2}\sum \limits_{i,j} (\sum \limits_{l} \gamma_{il} \gamma_{jl}) K(\mathbf{x}_i,\mathbf{x}_j)-  \sum \limits_{i,l} \gamma_{il}[\sum \limits_{j} \alpha_{jl}^t K(\mathbf{x}_i,\mathbf{x}_j)+e_{il}] && \nonumber \\
&= -\frac{1}{2}\sum \limits_{i,j} (\sum \limits_{l} \gamma_{il} \gamma_{jl}) K(\mathbf{x}_i,\mathbf{x}_j)-  \sum \limits_{l}\alpha_{tl}(\sum \limits_{j} \alpha_{jl}^t K(\mathbf{x}_j,\mathbf{x}_t) )+ \alpha_{ty_t}&& 
\end{flalign}
The last equality follows from assumption 2 and the construction of $\boldsymbol\gamma$ in \eqref{eq10}. 

As the \textit{second} step, we construct a feasible solution for the leave-one-out formulation \eqref{eq9} using the optimal solution for \eqref{eq5}. i.e., construct $\boldsymbol\alpha - \boldsymbol\beta$  as shown below,
\begin{flalign} \label{eq12}
\quad \alpha_{il} - \beta_{il} \leq C_i; \quad & \quad \forall\ (i,l) \in \lbrace (i,l) |\ \alpha_{il} < C_i;\ l = y_i \rbrace := A_1 && \nonumber \\
\quad \alpha_{il} - \beta_{il} \leq 0 ;\quad & \quad \forall\ (i,l) \in \lbrace (i,l) |\ \alpha_{il} < 0;\ l \neq y_i \rbrace := A_2 && \nonumber \\
\quad \sum \limits_{l} \beta{il} =0 ;\quad \quad &&& \nonumber
\end{flalign}
and,
\begin{flalign}
\quad \beta_{il} =0  \quad  \quad & \quad \forall (i,l) \notin SV_1-\{t\}&[\text{with} \ SV_1 = A_1 \cup A_2\ = \{ i \ | 0< \alpha_{iy_i} < C_i \} ] && \\
\quad \beta_{tl} =\alpha_{tl} \quad & \quad \forall l &(under Assumption 2)&& \nonumber
\end{flalign}
such that, it is a feasible solution for \eqref{eq9}. As before, define	
\begin{flalign} \label{eq13}
I_2 &= W(\boldsymbol\alpha)-W(\boldsymbol\alpha - \boldsymbol\beta) && \nonumber\\
&= -\frac{1}{2} \sum \limits_{i,j} \sum \limits_k \alpha_{il} \alpha_{jl} K(\mathbf{x}_i,\mathbf{x}_j) - \sum \limits_i \sum \limits_l \alpha_{il} e_{il}&& \nonumber \\
&\quad +\frac{1}{2}\sum \limits_{i,j} \sum \limits_{l}(\alpha_{il} - \beta_{il})(\alpha_{jl} - \beta_{jl})K(\mathbf{x}_i,\mathbf{x}_j)+\sum \limits_{i} \sum \limits_l (\alpha_{il} - \beta_{il})e_{il} &&\nonumber\\
&= \frac{1}{2}\sum \limits_{i,j} (\sum \limits_{l} \beta_{il} \beta_{jl}) K(\mathbf{x}_i,\mathbf{x}_j)-  \sum \limits_{i,j} (\sum \limits_{l}\beta_{il} \alpha_{jl})K(\mathbf{x}_i,\mathbf{x}_j)-\sum \limits_{i}\sum \limits_{l}\beta_{il}e_{il} && \nonumber \\
&= \frac{1}{2}\sum \limits_{i,j} (\sum \limits_{l} \beta_{il} \beta_{jl}) K(\mathbf{x}_i,\mathbf{x}_j)-  \sum \limits_{i,l} \beta_{il}[\sum \limits_{j} \alpha_{jl} K(\mathbf{x}_i,\mathbf{x}_j)+e_{il}] && \nonumber \\
&= \frac{1}{2}\sum \limits_{i,j} (\sum \limits_{l} \beta_{il} \beta_{jl}) K(\mathbf{x}_i,\mathbf{x}_j)-  \sum \limits_{l}\alpha_{tl}(\sum \limits_{j} \alpha_{jl} K(\mathbf{x}_j,\mathbf{x}_t) )+ \alpha_{ty_t}&& 
\end{flalign}
The last equality follows from assumption 2 and the construction of $\boldsymbol\beta$ in \eqref{eq12}. Moreover, from assumption 1, $\boldsymbol\beta =\boldsymbol\alpha - \boldsymbol\alpha^t=\boldsymbol\gamma$ satisfies the constraints in \eqref{eq9} and \eqref{eq11}. Hence for such a $\boldsymbol\beta,\boldsymbol\gamma$:\newline
\begin{flalign}\label{eq14}
&I_1 = I_2 =W(\boldsymbol\alpha)-W(\boldsymbol\alpha^t) &&&  \\
\Rightarrow &\sum \limits_{i,j} (\sum \limits_{l} \beta_{il} \beta_{jl}) K(\mathbf{x}_i,\mathbf{x}_j) &=\sum \limits_{l}\alpha_{tl}(\sum \limits_{j} \alpha_{jl} K(\mathbf{x}_j,\mathbf{x}_t) ) - \sum \limits_{l}\alpha_{tl}(\sum \limits_{j} \alpha_{jl}^t K(\mathbf{x}_j,\mathbf{x}_t))&& \nonumber \\
&&=\sum \limits_{l}\alpha_{tl}(\sum \limits_{j} \alpha_{jl} K(\mathbf{x}_j,\mathbf{x}_t) ) - (\alpha_{tl}\mathbf{g}_k)^{\top}(\sum \limits_{j} \alpha_{jl}^t K(\mathbf{x}_j,\mathbf{x}_t))&& \nonumber
\end{flalign}
The last equality follows from assumption 2, where $\mathbf{g}_k =[0,\ldots \underset{l^{th}}{1},\ldots,\underset{k^{th}}{-1},\ldots,0]$ for $\boldsymbol\alpha_{t} =[0,\ldots \underset{l^{th}=y_t}{\alpha_{tl}},\ldots,\underset{k^{th}}{-\alpha_{tl}},\ldots,0]$ (i.e. only two active elements for the support vector). \newline 
As the third and final step define, 
\begin{flalign} \label{eq15}
S_t^2 = &\quad \underset{\boldsymbol\beta}{\text{min}}\quad  \sum \limits_{i,j}(\sum \limits_{l}\beta_{il}\beta_{jl})K(\mathbf{x}_i,\mathbf{x}_j) &&\\
s.t. & \quad \alpha_{il} - \beta_{il} \leq C_i; \quad (i,l) \in A_1 - \{t\}&&\nonumber\\
& \quad \alpha_{il} - \beta_{il} \leq 0; \quad (i,l) \in A_2 - \{t\}&&\nonumber\\
&\quad \beta_{il} =0 ; \quad \forall (i,l) \notin SV_1-\{t\} && \nonumber\\
&\quad \beta_{tl} =\alpha_{tl} ;\quad \forall l && \nonumber\\
&\quad \sum \limits_{l} \beta_{il} =0 &&\nonumber
\end{flalign}
and let $\boldsymbol\beta^{\prime}$ be the minimizer for \eqref{eq15}. Then,
\begin{flalign}
&W(\boldsymbol\alpha^t) \geq W(\boldsymbol\alpha-\boldsymbol\beta^{\prime}) \quad \text{[From \eqref{eq9}]} &&\nonumber\\
\Rightarrow & W(\boldsymbol\alpha) - W(\boldsymbol\alpha^t) \leq W(\boldsymbol\alpha) - W(\boldsymbol\alpha-\boldsymbol\beta^{\prime}) && \nonumber\\
\Rightarrow & \sum \limits_{i,j}(\sum \limits_{l}\beta_{il}\beta_{jl})K(\mathbf{x}_i,\mathbf{x}_j) \quad \leq \quad S_t^2 && \nonumber
\end{flalign} 

From \textit{Assumption 1}, $(\boldsymbol\alpha -\boldsymbol\alpha^t)$ is a feasible solution for \eqref{eq15} which gives : $ S_t^2 \leq \sum \limits_{i,j}(\sum \limits_{l}\beta_{il}\beta_{jl})K(\mathbf{x}_i,\mathbf{x}_j) $. Combining the above inequality, $S_t^2 = \sum \limits_{i,j}(\sum \limits_{l}\beta_{il}\beta_{jl})K(\mathbf{x}_i,\mathbf{x}_j)$.
Moreover, under \textit{Assumption 1} the inequality constraints in \eqref{eq15} are not activated. Hence,  
$S_t^2 =\{\underset{\mathbf{\beta}}{\text{min}}\ \sum \limits_{i,j}(\sum \limits_{l}\beta_{il}\beta_{jl})K(\mathbf{x}_i,\mathbf{x}_j)|\ \boldsymbol\beta_{t} =\boldsymbol\alpha_{t} ;\ \sum \limits_{l} \beta_{il} =0\ ; (i,j)\in SV_1\}$.  \quad \textbf{Proved}.
\newline
\subsection{Proof of Corollary 1}
The Span is defined as:
\begin{flalign} \label{eq16}
S_t^2 &= \underset{\mathbf{\boldsymbol\beta}}{\text{min}} \; \sum \limits_{i,j}(\sum \limits_{l} \beta_{il}\beta_{jl})K(\mathbf{x}_i,\mathbf{x}_j) &&  \\ 
& \quad  s.t. \quad \quad \beta_{tl} = \alpha_{tl} \quad ; \quad \forall l = 1, \ldots, L && \nonumber \\
& \quad \quad \quad \quad \sum\limits_{l} \beta_{il} = 0 \quad ; \quad \forall (i,j) \in SV_1 && \nonumber
\end{flalign}

\begin{flalign}
&= \underset{\boldsymbol\beta}{\text{min}} \; \sum \limits_{l} (\alpha_{tl}\alpha_{tl}) K(\mathbf{x}_t,\mathbf{x}_t) + 2 \sum\limits_{i\neq t}\sum \limits_l \alpha_{tl}\beta_{il}K(\mathbf{x}_t,\mathbf{x}_i) + \sum \limits_{(i,j) \neq t}(\sum \limits_{l} \beta_{il}\beta_{jl})K(\mathbf{x}_i,\mathbf{x}_j) && \nonumber \\
& \quad s.t. \quad \underset{\mathbf{A}}{\underbrace{(\mathbf{I}_{|SV_1 - \{t\}|}\otimes \mathbf{1}_L)}} \boldsymbol\beta = \mathbf{0} && \nonumber \\
&=\; \underset{\boldsymbol\beta}{\text{min}}\; \underset{\boldsymbol\mu}{\text{max}}\; \boldsymbol\alpha_t^{\top} [K(\mathbf{x}_t,\mathbf{x}_t) \otimes \mathbf{I}_L] \boldsymbol\alpha_t + 2 \sum\limits_{i\neq t}\sum \limits_l \alpha_{tl}\beta_{il}K(\mathbf{x}_t,\mathbf{x}_i) + \sum \limits_{(i,j) \neq t}(\sum \limits_{l} \beta_{il}\beta_{jl})K(\mathbf{x}_i,\mathbf{x}_j) &&\nonumber\\
& \quad \quad \quad \quad  \quad + 2 \boldsymbol\mu^{\top} \mathbf{A} \boldsymbol\beta \quad \quad \quad \quad \quad \quad \quad \quad \quad \quad \quad \quad \quad \quad \quad \quad \quad \quad  (\boldsymbol\mu := \text{Lagrange Multiplier})&&\nonumber\\
&= \boldsymbol\alpha_t^{\top} [K(\mathbf{x}_t,\mathbf{x}_t) \otimes \mathbf{I}_L] \boldsymbol\alpha_t \; + \; \underset{\boldsymbol\beta}{\text{min}}\; \underset{\boldsymbol\mu}{\text{max}}\; \underset{L(\lambda)}{\underbrace{2\boldsymbol\alpha_t^{\top}(\mathbf{H}_t^{(-t)})^{\top}\boldsymbol\lambda + \boldsymbol\lambda \mathbf{H}^{(-t)}\boldsymbol\lambda}} \quad \quad \quad \quad (\text{with}\quad \boldsymbol\lambda=[\boldsymbol\beta;\boldsymbol\mu]) && \nonumber
\end{flalign}
where,\quad  $\mathbf{I}_{|SV_1 - \{t\}|}:=$ Identity Matrix of size $|SV_1 - \{t\}|$, \newline $\mathbf{H}^{(-\mathbf{t})} := \; (t-1)L+1,\ldots, tL \; \text{rows/columns of matrix}\; \mathbf{H} $ (in \eqref{eq7}) removed; and \newline $\mathbf{H}_{\mathbf{t}}^{(-\mathbf{t})}:= (t-1)L+1 ,\ldots, tL \; \text{columns of} \; \mathbf{H}$.  \newline Further, at saddle point :  $\bigtriangledown_{\boldsymbol\lambda}L(\boldsymbol\lambda) = 0 \quad \Rightarrow \boldsymbol\lambda^{*} = - [\mathbf{H}^{(-\mathbf{t})}]^{-1} \mathbf{H}_{\mathbf{t}}^{(-\mathbf{t})} \boldsymbol\alpha_t$. \newline Hence, 
\begin{flalign} \label{eq17}
S_t^2 &= \boldsymbol\alpha_t^{\top} [(K(\mathbf{x}_t,\mathbf{x}_t)\otimes\mathbf{I}_L)-(\mathbf{H}_{\mathbf{t}}^{(-\mathbf{t})})^{\top} (\mathbf{H}^{(-\mathbf{t})})^{-1} \mathbf{H}_{\mathbf{t}}^{(-\mathbf{t})}]\boldsymbol\alpha_t && \nonumber \\
&= \boldsymbol\alpha_t^{\top} (\mathbf{H}^{-1})_{\mathbf{tt}}\boldsymbol\alpha_t && 
\end{flalign} 
where, $(\mathbf{H}^{-1})_{\mathbf{tt}} \quad := \text{sub-matrix of } \;  \mathbf{H}^{-1} \ \text{for index }\;   i\  = (t-1)L+1,\ldots ,tL $. \quad \textbf{Proved}.
\newline
\subsection{Proof of Corollary 2}
The proof has three steps and mainly depends on the contribution of a sample to the leave-one-out error. 
\begin{itemize}
\item[--] \textit{First}, for a sample $(\mathbf{x}_t,y_t)$ which is not a support vector, i.e. $t \notin SV$ and $t \in \mathcal{T}$ (Training set); it lies outside margin borders. Dropping such a sample does not change the original solution \eqref{eq5}. Hence, it does not contribute to an error.
\item[--] \textit{Secondly}, for a sample $(\mathbf{x}_t,y_t)$ with $t \in SV_1$ and $t \in \mathcal{T}$ (Training set) Theorem 1 holds. For a leave-one-out error, $(\alpha_{tl}\mathbf{g}_k)^{\top} \sum \limits_{i \in SV^t} \sum \limits_{l} \alpha_{il}^t K(\mathbf{x}_i,\mathbf{x}_t) \leq 0 \Rightarrow \boldsymbol\alpha_t^{\top} [(\mathbf{H}^{-1})_{\mathbf{tt}}]^{-1} \boldsymbol\alpha_t \geq  \boldsymbol\alpha_t^{\top} \sum\limits_{i \in SV} \sum\limits_{l} \alpha_{il} K(\mathbf{x}_i,\mathbf{x}_t)$. [From \eqref{eq14}]
\item[--] \textit{Finally}, for a sample $(\mathbf{x}_t,y_t)$ with $t \in SV_2$ and $t \in \mathcal{T}$ (Training set) we add to the leave-one-out error.

\end{itemize}

\quad \textbf{Proved}.

\section{Additional Results}
\subsection{Histogram of projections}
\subsubsection{GTSRB dataset}

\begin{figure}[!htb]
\centering
\includegraphics[height=2cm, width=10cm]{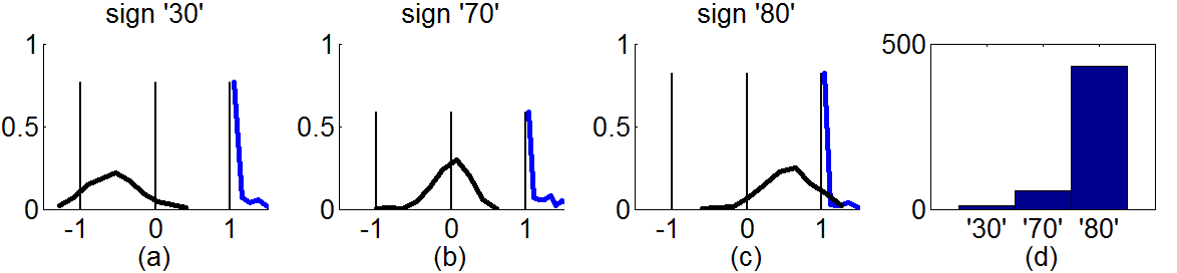}
\caption{Typical histogram of projection of training samples (shown in \textcolor{blue}{\textbf{blue}}) and universum samples (shown in \textbf{black}) onto the multiclass SVM model (with $C=1$). Decision functions for (a) sign `30'. (b) sign `70'.(c) sign `80'. (e) frequency plot of predicted labels for universum samples(`\textit{roadworks}').} \label{histSVM2}
\includegraphics[height=2cm, width=10cm]{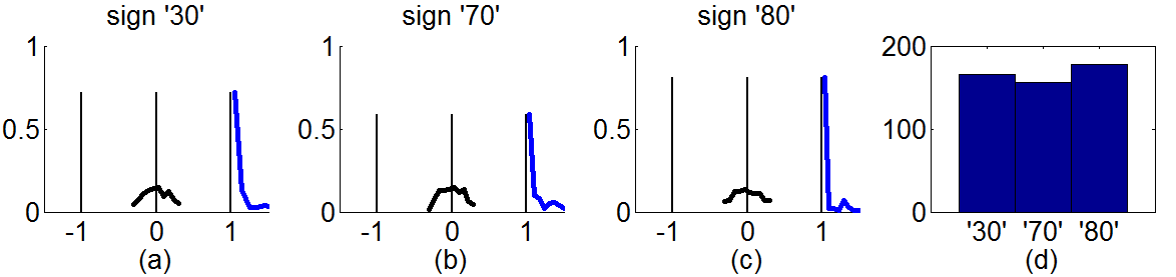}
\caption{Typical histogram of projection of training samples (shown in \textcolor{blue}{\textbf{blue}}) and universum samples (shown in \textbf{black}) onto the multiclass U-SVM model (with $\Delta=0.5$). Decision functions for (a) sign `30'. (b) sign `70'.(c) sign `80'. (e) frequency plot of predicted labels for universum samples (`\textit{roadworks}').} \label{histUSVM2}
\end{figure}

Figs \ref{histSVM2} and \ref{histUSVM2} provide the histograms and the frequency plots for SVM/MU-SVM models for GTSRB dataset using sign `\textit{roadworks}' (as universum). As seen from Fig. \ref{histSVM2}, the optimal SVM model has high separability for the training samples. Here, the universum samples are biased towards the positive side of the decision boundary of the sign `80' (see Fig \ref{histSVM2}(c)) and hence predominantly gets classified as sign `80'(see Fig.\ref{histSVM2} (d)). As seen from Figs  \ref{histUSVM2} (a)-(c), applying the MU-SVM model preserves the separability of the training samples and additionally reduces the spread of the universum samples. For such a model the uncertainity due to universum samples is uniform across all the classes i.e. signs `30',`70' and `80' (see Fig. \ref{histUSVM2}(d)). The resulting MU-SVM model has higher contradiction on the universum samples and provides better generalization in comparison to SVM (see Table. 2).

\subsubsection{ABCDETC dataset}

\begin{figure}[!htb]
\includegraphics[height=2cm, width=12cm]{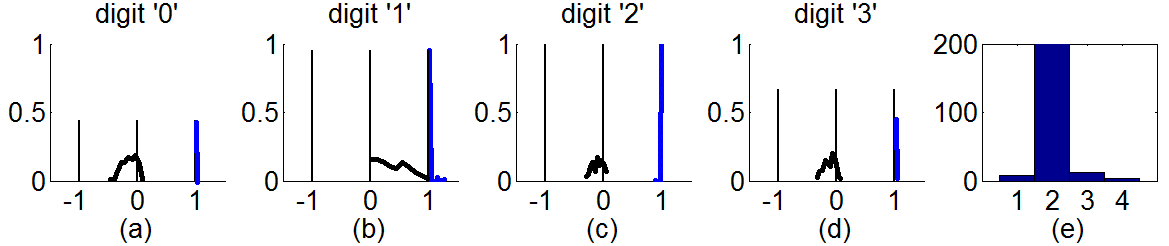}
\caption{Typical histogram of projection of training samples (in \textcolor{blue}{\textbf{blue}}) and universum samples (in \textbf{black}) onto the SVM model( with $C=1$ and $\gamma = 2^{-7} $ ).(a) digit `0'. (b) digit `1'.(c) digit `2'. (d) digit `3'. (e) frequency plot of predicted labels for universum samples (lowercase letter `i').} \label{histSVMi}
\includegraphics[height=2cm, width=12cm]{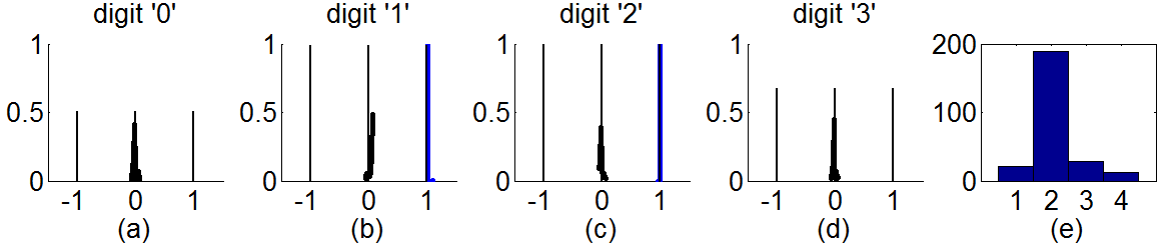}
\caption{Typical histogram of projection of training samples (in \textcolor{blue}{\textbf{blue}}) and universum samples (in \textbf{black}) onto the U-SVM model (with $C^{*}/C=0.6$ and $\Delta = 0$ ).(a) digit `0'. (b) digit `1'.(c) digit `2'. (d) digit `3'.(e) frequency plot of predicted class labels for universum samples (lowercase letter `i').} \label{histUSVMi}
\end{figure}

Here we present the histograms and the frequency plots for SVM/MU-SVM models for ABCDETC dataset using letter `\textit{i}' (as universum). Here, the SVM model results in a wide distribution of the universum samples (see Fig. \ref{histSVMi} (a)-(d)) and predicts majority of the universum samples as digit '1'. Applying MU-SVM results in a narrower distribution of the universum samples (see Fig \ref{histUSVMi} (a)-(d)) and hence a more random prediction on the universum samples (see Fig \ref{histUSVMi}). This results in a more generalizable model in comparison to SVM (see Table 2)

\subsection{Experiments with varying Universum size}
\begin{table}[!htb] 
\centering
\caption{Comparison of average test error for different Universa with increase in Universum samples. The results show mean test error in \%, over 10 runs. The numbers in parentheses denote the standard deviations.} \label{tab}
\begin{tabular}{|c||c|c|c|c|} \hline 
\specialcell{GTSRB\\ (dataset)} & \multicolumn{4}{|c|}{ Training size = 300 (100 per class), \quad Test size = 1500 (500 per class) }\\ \hline
Universum size & \quad m = 250 \quad & \quad m = 500 \quad & \quad m = 750 \quad & \quad m = 1000 \quad \\ 
\hline
SVM& \specialcell{7.23\\(1.02)}& - & - & - \\
\hline
\specialcell{MU-SVM\\(no-entry)}& \specialcell{6.93\\(0.98)}& \specialcell{6.48\\(0.52)} & \specialcell{6.43\\(0.59)} & \specialcell{6.41\\(0.6)} \\
\hline
\specialcell{MU-SVM\\(roadworks)}& \specialcell{6.83\\(0.75)}& \specialcell{6.7\\(0.68)} & \specialcell{6.72\\(0.51)} & \specialcell{6.68\\(0.71)} \\
\hline 
\end{tabular}
\end{table}

This set of experiment demonstrates how the generalization performance of MU-SVM is affected by the number of Universum data samples for the GTSRB dataset. Here we use the same setting as provided in Table 1, except we vary the number of universum samples as shown in Table 4. Table 4 shows the performance comparison between multiclass SVM vs. MU-SVM, suggesting that, for both types of Universa,  prediction performance of MU-SVM improves with the number universum samples. However,increasing the number of universum samples above certain value ($\sim$ 500) does not provide additional improvement.